\documentclass[runningheads]{llncs}
\usepackage{graphicx}
\usepackage{comment}
\usepackage{amsmath,amssymb} 
\usepackage{color}

\usepackage{subfig}
\usepackage{booktabs}
\usepackage{wrapfig}

\usepackage[pagebackref=true,breaklinks=true,letterpaper=true,colorlinks,bookmarks=false]{hyperref}

\def\eg{\emph{e.g.}}
\def\ie{\emph{i.e.}}

\newcommand{\zongxin}[1]{#1}

\begin{document}
\pagestyle{headings}
\mainmatter
\def\ECCVSubNumber{3385}  

\title{Collaborative Video Object Segmentation by Foreground-Background Integration} 

\titlerunning{CFBI: Collaborative VOS by Foreground-Background Integration}
%
\author{Zongxin Yang\inst{1*,2} \and
Yunchao Wei\inst{2} \and
Yi Yang\inst{2}}
\authorrunning{Z. Yang et al.}
%
\institute{Baidu Research \and
ReLER, Centre for Artificial Intelligence, University of Technology Sydney\\
\email{zongxin.yang@student.uts.edu.au \{yunchao.wei,yi.yang\}@uts.edu.au}}
\maketitle

\renewcommand{\thefootnote}{*}
\footnotetext{This work was done when Zongxin Yang interned at Baidu Research. Yi Yang is the corresponding author.}

\setlength{\intextsep}{0pt}

\begin{abstract}
This paper investigates the principles of embedding learning to tackle the challenging semi-supervised video object segmentation.
Different from previous practices that only explore the embedding learning using pixels from foreground object (s), we consider background should be equally treated and thus propose Collaborative video object segmentation by Foreground-Background Integration (CFBI) approach. Our CFBI implicitly imposes the feature embedding from the target foreground object and its corresponding background to be contrastive, promoting the segmentation results accordingly.  With the feature embedding from both foreground and background, our CFBI performs the matching process between the reference and the predicted sequence from both pixel and instance levels,  making the CFBI be robust to various object scales. We conduct extensive experiments on three popular benchmarks, \ie, DAVIS 2016, DAVIS 2017, and YouTube-VOS. Our CFBI achieves the performance ($\mathcal{J}$\&$\mathcal{F}$) of 89.4\%, 81.9\%, and 81.4\%, respectively, outperforming all the other state-of-the-art methods. Code: \url{https://github.com/z-x-yang/CFBI}.


\end{abstract}

\section{Introduction}

Video Object Segmentation (VOS) is a fundamental task in computer vision with many potential applications, including augmented reality~\cite{ngan2011video} and self-driving cars~\cite{zhang2016instance}. In this paper, we focus on semi-supervised VOS, which targets on segmenting a particular object across the entire video sequence based on the object mask given at the first frame. 
\zongxin{The development of semi-supervised VOS can benefit many related tasks, such as video instance segmentation~\cite{vis,Feng_2019_ICCV} and interactive video object segmentation~\cite{oh2019fast,miao2020memory,liangmemory}.}

Early VOS works~\cite{osvos,onavos,premvos} rely on fine-tuning with the first frame in evaluation, which heavily slows down the inference speed. Recent works (\eg,~\cite{osmn,feelvos,spacetime}) aim to avoid fine-tuning and achieve better run-time. 
In these works, STMVOS~\cite{spacetime} introduces memory networks to learn to read sequence information and outperforms all the fine-tuning based methods. However, STMVOS relies on simulating extensive frame sequences using large image datasets~\cite{voc,coco,cheng2014global,shi2015hierarchical,semantic} for training. The simulated data significantly boosts the performance of STMVOS but makes the training procedure elaborate. Without simulated data, FEELVOS~\cite{feelvos} adopts a semantic pixel-wise embedding together with a global (between the first and current frames) and a local (between the previous and current frames) matching mechanism to guide the prediction. The matching mechanism is simple and fast, but the performance is not comparable with STMVOS.

\begin{wrapfigure}[17]{R}{0.55\textwidth}
\centering 

\includegraphics[width=0.98\linewidth]{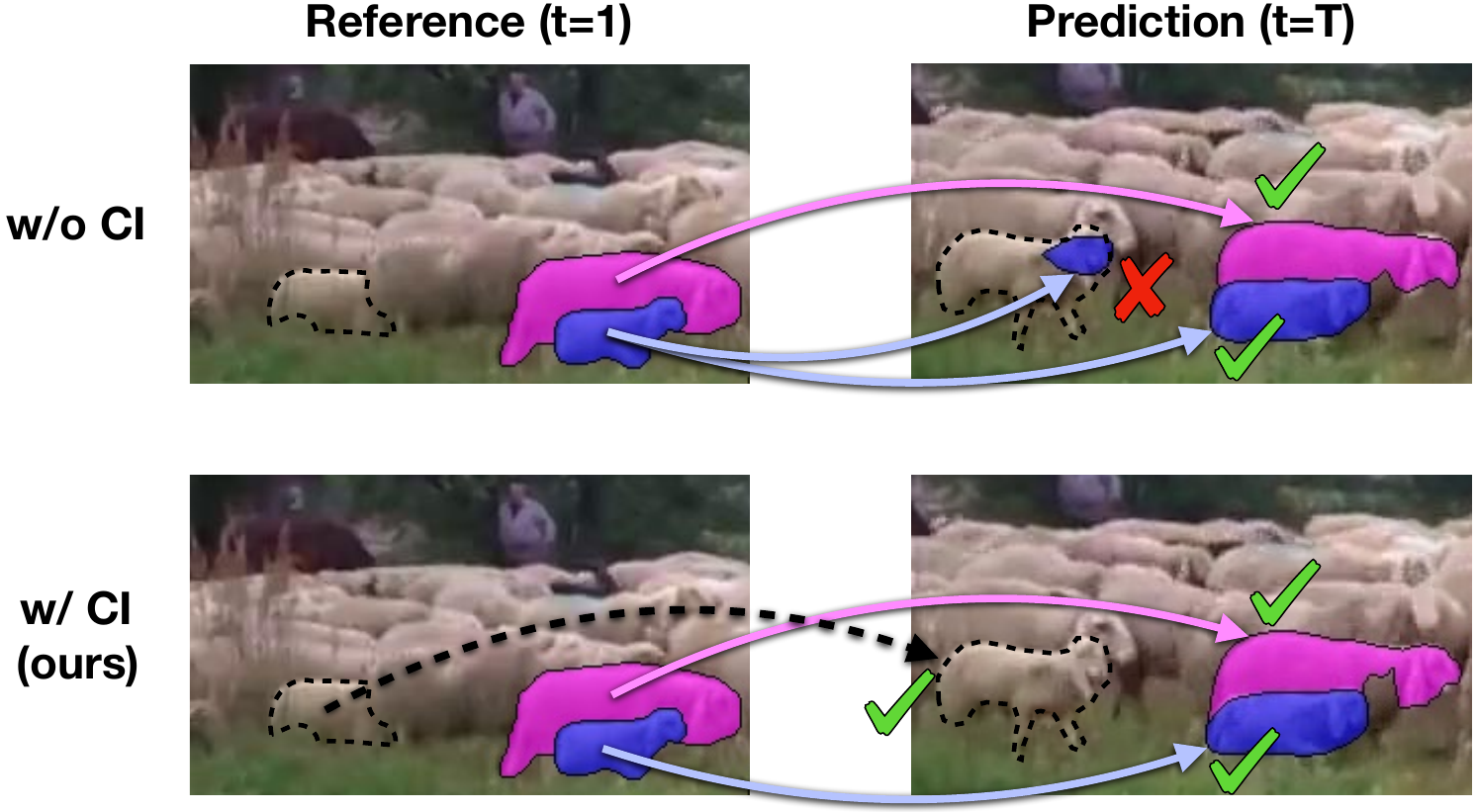}

\caption{CI means collaborative integration. There are two foreground sheep (pink and blue). In the top line, the contempt of background matching leads to a confusion of sheep's prediction. In the bottom line, we relieve the confusion problem by introducing background matching (dot-line arrow).}\label{fig:cl}

\end{wrapfigure}

Even though the efforts mentioned above have made significant progress, current state-of-the-art works pay little attention to the feature embedding of background region in videos and only focus on exploring robust matching strategies for the foreground object (s). Intuitively, it is easy to extract the foreground region from a video when precisely removing all the background. Moreover, modern video scenes commonly focus on many similar objects, such as the cars in car racing, the people in a conference, and the animals on a farm. For these cases, the contempt of integrating foreground and background embeddings traps VOS in an unexpected background confusion problem. As shown in Fig.~\ref{fig:cl}, if we focus on only the foreground matching like FEELVOS, a similar and same kind of object (sheep here) in the background is easy to confuse the prediction of the foreground object. Such an observation motivates us that the background should be equally treated compared with the foreground so that better feature embedding can be learned to relieve the background confusion and promote the accuracy of VOS.

We propose a novel framework for Collaborative video object segmentation by Foreground-Background Integration (CFBI) based on the above motivation. 
Different from the above methods, we not only extract the embedding and do match for the foreground target in the reference frame, but also for the background region to relieve the background confusion.
Besides, our framework extracts two types of embedding (\ie, pixel-level, and instance-level embedding) for each video frame to cover different scales of features. Like FEELVOS, we employ pixel-level embedding to match all the objects' details with the same global \& local mechanism. However, the pixel-level matching is not sufficient and robust to match those objects with larger scales and may bring unexpected noises due to the pixel-wise diversity. Thus we introduce instance-level embedding to help the segmentation of large-scale objects by using attention mechanisms.
Moreover, we propose a collaborative ensembler to aggregate the foreground \& background and pixel-level \& instance-level information and learn the collaborative relationship among them implicitly.
\zongxin{For better convergence, we take a balanced random-crop scheme in training to avoid learned attributes being biased to the background attributes.}
All these proposed strategies can significantly improve the quality of the learned collaborative embeddings for conducting VOS while keeping the network simple yet effective simultaneously.

We perform extensive experiments on DAVIS~\cite{davis2016,davis2017}, and YouTube-VOS~\cite{youtubevos} to validate the effectiveness of the proposed CFBI approach. Without any bells and whistles (such as the use of simulated data, fine-tuning or post-processing), CFBI outperforms all other state-of-the-art methods on the validation splits of DAVIS 2016 (ours, $\mathcal{J}\&\mathcal{F}$ $\mathbf{89.4\%}$), DAVIS 2017 ($\mathbf{81.9\%}$) and YouTube-VOS ($\mathbf{81.4\%}$) while keeping a competitive single-object inference speed of about 5 FPS. By additionally applying multi-scale \& flip augmentation at the testing stage, the accuracy can be further boosted to $\mathbf{90.1\%}$, $\mathbf{83.3\%}$ and $\mathbf{82.7\%}$, respectively. 
We hope our simple yet effective CFBI will serve as a solid baseline and help ease VOS's future research.

\section{Related Work}

\noindent\textbf{Semi-supervised Video Object Segmentation.}
Many previous methods for semi-supervised VOS rely on fine-tuning at test time. Among them, OSVOS~\cite{osvos} and MoNet~\cite{xiao2018monet} fine-tune the network on the first-frame ground-truth at test time. OnAVOS~\cite{onavos} extends the first-frame fine-tuning by an online adaptation mechanism, \ie, online fine-tuning. MaskTrack~\cite{masktrack} uses optical flow to propagate the segmentation mask from one frame to the next. PReMVOS~\cite{premvos} combines four different neural networks (including an optical flow network~\cite{flownet}) using extensive fine-tuning and a merging algorithm. Despite achieving promising results, all these methods are seriously slowed down by fine-tuning during inference.

Some other recent works (\eg,~\cite{osmn,favos}) aim to avoid fine-tuning and achieve a better run-time. OSMN~\cite{osmn} employs two networks to extract the instance-level information and make segmentation predictions, respectively.  PML~\cite{pml} learns a pixel-wise embedding with the nearest neighbor classifier. Similar to PML, VideoMatch~\cite{videomatch} uses a soft matching layer that maps the pixels of the current frame to the first frame in a learned embedding space. Following PML and VideoMatch, FEELVOS~\cite{feelvos} extends the pixel-level matching mechanism by additionally matching between the current frame and the previous frame. Compared to the methods with fine-tuning, FEELVOS achieves a much higher speed, but there is still a gap inaccuracy. Like FEELVOS, RGMP~\cite{rgmp} and STMVOS~\cite{spacetime} does not require any fine-tuning. STMVOS, which leverages a memory network to store and read the information from past frames, outperforms all the previous methods. However, STMVOS relies on an elaborate training procedure using extensive simulated data generated from multiple datasets. Moreover, the above methods do not focus on background matching.

Our CFBI utilizes both the pixel-level and instance-level embeddings to guide prediction. Furthermore, we propose a collaborative integration method by additionally learning background embedding. 

\noindent\textbf{Attention Mechanisms.}
Recent works introduce the attention mechanism into convolutional networks (\eg, ~\cite{attention_conv1,attention_conv2}). 
Following them, SE-Nets~\cite{senet} introduced a lightweight gating mechanism that focuses on enhancing the representational power of the convolutional network by modeling channel attention. Inspired by SE-Nets, CFBI uses an instance-level average pooling method to embed collaborative instance information from pixel-level embeddings. After that, we conduct a channel-wise attention mechanism to help guide prediction. Compared to OSMN, which employs an additional convolutional network to extract instance-level embedding, our instance-level attention method is more efficient and lightweight.

\begin{figure}[t!]
    \centering
    \includegraphics[width=0.9\linewidth]{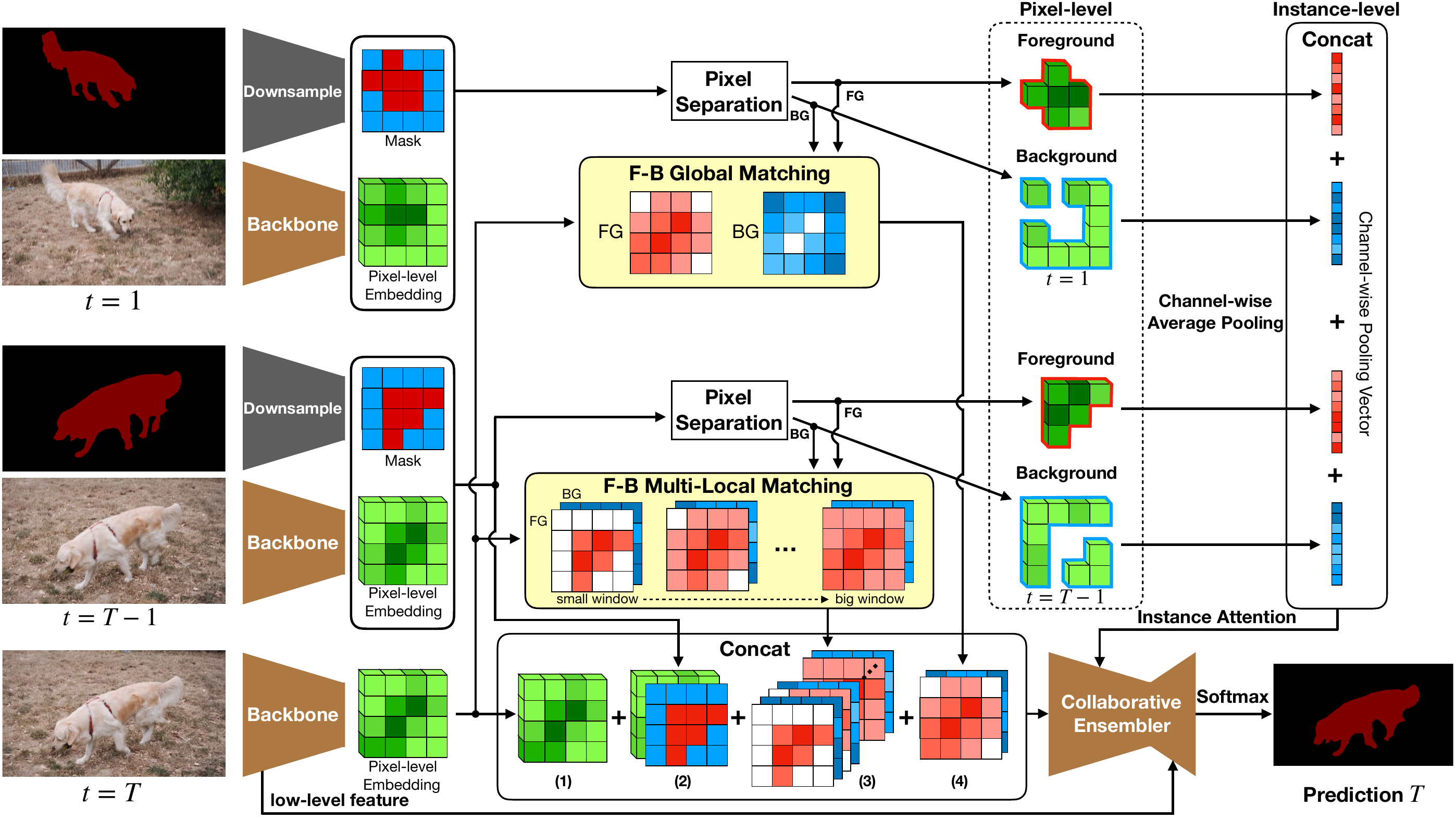}
    \caption{An \textbf{overview} of CFBI. F-G denotes Foreground-Background. We use \textcolor{red}{red} and \textcolor{blue}{blue} to indicate foreground and background separately. The deeper the red or blue color, the higher the confidence. Given the first frame ($t=1$), previous frame ($t=T-1$), and current frame ($t=T$), we firstly extract their pixel-wise embedding by using a backbone network. Second, we separate the first and previous frame embeddings into the foreground and background pixels based on their masks. After that, we use F-G pixel-level matching and instance-level attention to guide our collaborative ensembler network to generate a prediction.}
    \label{fig:overview}

\end{figure}

\section{Method}\label{sec:model}

\noindent\textbf{Overview.} Learning foreground feature embedding has been well explored by previous practices (\eg,~\cite{osmn,feelvos}). OSMN proposed to conduct an instance-level matching, but such a matching scheme fails to consider the feature diversity among the details of the target's appearance and results in coarse predictions. PML and FEELVOS alternatively adopt the pixel-level matching by matching each pixel of the target, which effectively takes the feature diversity into account and achieves promising performance. Nevertheless, performing pixel-level matching may bring unexpected noises in the case of some pixels from the background are with a similar appearance to the ones from the foreground (Fig.~\ref{fig:cl}).

To overcome the problems raised by the above methods and promote the foreground objects from the background, we present Collaborative video object segmentation by Foreground-Background Integration (CFBI), as shown in Figure~\ref{fig:overview}. We use red and blue to indicate foreground and background separately. First, beyond learning feature embedding from foreground pixels, our CFBI also considers embedding learning from background pixels for collaboration. Such a learning scheme will encourage the feature embedding from the target object and its corresponding background to be contrastive, promoting the segmentation results accordingly. Second, we further conduct the embedding matching from both pixel-level and instance-level with the collaboration of pixels from the foreground and background. For the pixel-level matching, 
we improve the robustness of the local matching under various object moving rates. For the instance-level matching, we design an instance-level attention mechanism to augment the pixel-level matching efficiently. Moreover, to implicitly aggregate the learned foreground \& background and pixel-level \& instance-level information, we employ a collaborative ensembler to construct large receptive fields and make precise predictions.

\subsection{Collaborative Pixel-level Matching}

For the pixel-level matching, we adopt a global and local matching mechanism similar to FEELVOS for introducing the guided information from the first and previous frames, respectively. Unlike previous methods~\cite{pml,feelvos}, we additionally incorporate background information and apply multiple windows in the local matching, which is shown in the middle of Fig.~\ref{fig:overview}. 

For incorporating background information, we firstly redesign the pixel distance of~\cite{feelvos} to further distinguish the foreground and background.
Let $B_t$ and $F_t$ denote the pixel sets of background and all the foreground objects of frame $t$, respectively. We define a new distance between pixel $p$ of the current frame $T$ and pixel $q$ of frame $t$ in terms of their corresponding embedding, $e_p$ and $e_q$, by
\begin{equation} \label{equ:distance}
    D_t(p,q)=
        \begin{cases}
            1-\frac{2}{1+exp(||e_p-e_q||^2+b_B)} & \text{if } q \in B_t\\
            1-\frac{2}{1+exp(||e_p-e_q||^2+b_F)} & \text{if } q \in F_t
        \end{cases},
\end{equation}
where $b_B$ and $b_F$ are trainable background bias and foreground bias. We introduce these two biases to make our model be able further to learn the difference between foreground distance and background distance.

\noindent\textbf{Foreground-Background Global Matching.} Let $\mathcal{P}_t$ denote the set of all pixels (with a stride of 4) at time $t$ and $\mathcal{P}_{t,o}\subseteq \mathcal{P}_{t}$ is the set of pixels at time $t$ which belongs to the foreground object $o$. The global foreground matching between one pixel $p$ of the current frame $T$ and the pixels of the first reference frame (\ie, $t=1$) is,
\begin{equation} \label{equ:global_f}
    G_{T,o}(p)=\min_{q\in\mathcal{P}_{1,o}} D_1(p,q).
\end{equation}
Similarly, let $\mathcal{\overline{P}}_{t,o} =\mathcal{P}_t \backslash \mathcal{P}_{t,o}$ denote the set of relative background pixels of object $o$ at time $t$, and the global background matching is,
\begin{equation} \label{equ:global_b}
    \overline{G}_{T,o}(p)=\min_{q\in\mathcal{\overline{P}}_{1,o}} D_{1}(p,q).
\end{equation}

\noindent\textbf{Foreground-Background Multi-Local Matching.}

\setlength{\intextsep}{-10pt}
\begin{wrapfigure}[20]{R}{0.37\textwidth}
\center

\subfloat[Slow moving rate]{
\label{fig:slow}
\includegraphics[width=0.9\linewidth]{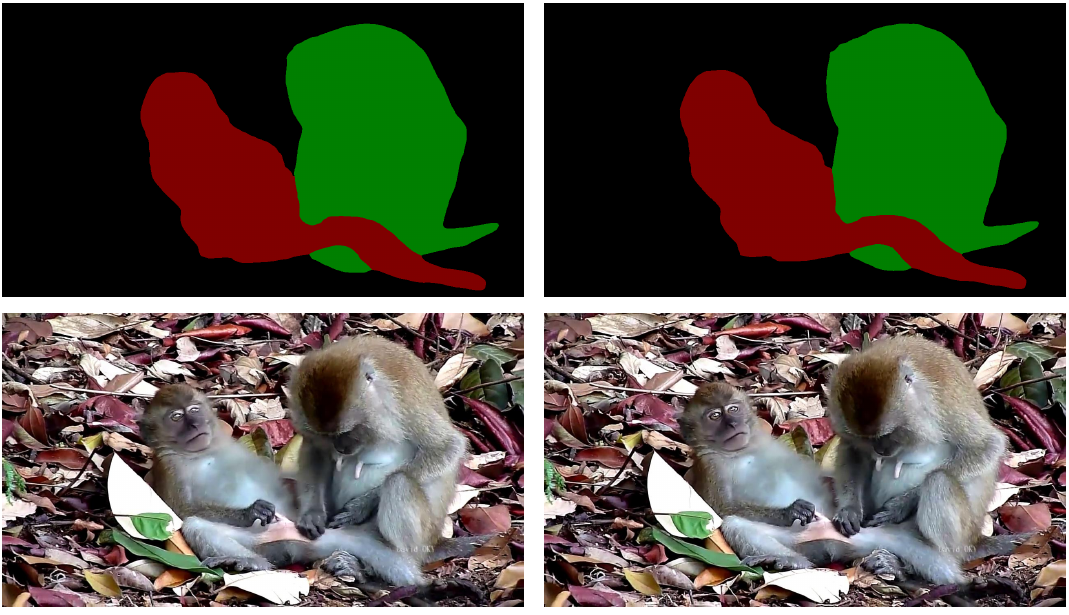}
}

\subfloat[Fast moving rate]{
\label{fig:fast}
\includegraphics[width=0.9\linewidth]{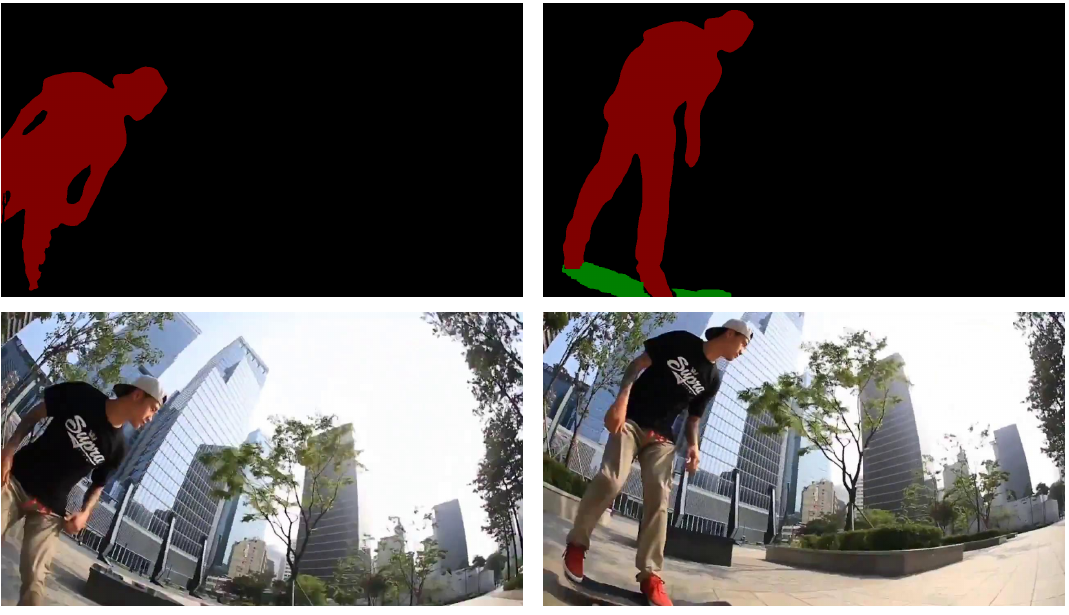}
}

\caption{The moving rate of objects across two adjacent frames is largely variable for different sequences. Examples are from YouTube-VOS~\cite{youtubevos}.}\label{fig:offset}
\end{wrapfigure}

\noindent In FEELVOS, the local matching is limited in only one fixed extent of neighboring pixels, but the offset of objects across two adjacent frames in VOS is variable, as shown in Fig.~\ref{fig:offset}. Thus, we propose to apply the local matching mechanism on different scales and let the network learn how to select an appropriate local scale, which makes our framework more robust to various moving rates of objects. Notably, we use the intermediate results of the local matching with the largest window to calculate on other windows. Thus, the increase of computational resources of our multi-local matching is negligible.

\setlength{\intextsep}{0pt}

Formally, let $K=\{k_1,k_2,...,k_n\}$ denote all the neighborhood sizes and $H(p,k)$ denote the neighborhood set of pixels that are at most $k$ pixels away from $p$ in both $x$ and $y$ directions, our foreground multi-local matching between the current frame $T$ and its previous frame $T-1$ is
\begin{equation} 
    ML_{T,o}(p,K)=\{L_{T,o}(p,k_1),L_{T,o}(p,k_2),...,L_{T,o}(p,k_n)\},
\end{equation}
where
\begin{equation} \label{equ:local_f}
    L_{T,o}(p,k)=
        \begin{cases}
            \min_{q\in\mathcal{P}^{p,k}_{T-1,o}} D_{T-1}(p,q) & \text{if }\mathcal{P}^{p,k}_{T-1,o}\neq\emptyset \\
            1 & \text{otherwise}
        \end{cases}.
\end{equation}
Here, $\mathcal{P}^{p,k}_{T-1,o}:=\mathcal{P}_{T-1,o}\cap H(p,k)$ denotes the pixels in the local window (or neighborhood). And our background multi-local matching is
\begin{equation} 
    \overline{ML}_{T,o}(p,K)=\{\overline{L}_{T,o}(p,k_1),\overline{L}_{T,o}(p,k_2),...,\overline{L}_{T,o}(p,k_n)\},
\end{equation}
where
\begin{equation} \label{equ:local_b}
    \overline{L}_{T,o}(p,k)=
        \begin{cases}
            \min_{q\in\mathcal{\overline{P}}_{T-1,o}^{p,k}} D_{T-1}(p,q) & \text{if }\mathcal{\overline{P}}_{T-1,o}^{p,k}\neq\emptyset \\
            1 & \text{otherwise}
        \end{cases}.
\end{equation}
Here similarly, $\mathcal{\overline{P}}^{p,k}_{T-1,o}:=\mathcal{\overline{P}}_{T-1,o}\cap H(p,k)$.

In addition to the global and multi-local matching maps, we concatenate the pixel-level embedding feature and mask of the previous frame with the current frame feature. FEELVOS demonstrates the effectiveness of concatenating the previous mask. Following this, we empirically find that introducing the previous embedding can further improve the performance ($\mathcal{J}$\&$\mathcal{F}$) by about $0.5\%$.

In summary, the output of our collaborative pixel-level matching is a concatenation of (1) the pixel-level embedding of the current frame, (2) the pixel-level embedding and mask of the previous frame, (3) the multi-local matching map and (4) the global matching map, as shown in the bottom box of Fig.~\ref{fig:overview}. 

\setlength{\intextsep}{-10pt}
\begin{wrapfigure}[22]{r}{0.3\textwidth}
\center

\includegraphics[width=0.98\linewidth]{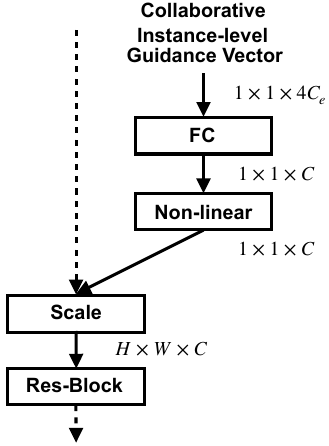}

\caption{The trainable part of the instance-level attention. $C_e$ denotes the channel dimension of pixel-wise embedding. $H$, $W$, $C$ denote the height, width, channel dimension of CE features.}
\label{fig:instance}

\end{wrapfigure}

\subsection{Collaborative Instance-level Attention}

As shown in the right of Fig~\ref{fig:overview}, we further design a Collaborative instance-level attention mechanism to guide the segmentation for large-scale objects. 

After getting the pixel-level embeddings of the first and previous frames, we separate them into foreground and background pixels (\ie, $\mathcal{P}_{1,o}$, $\mathcal{\overline{P}}_{1,o}$, $\mathcal{P}_{T-1,o}$, and $\mathcal{\overline{P}}_{T-1,o}$) according to their masks. Then, we apply channel-wise average pooling on each group of pixels to generate a total of four instance-level embedding vectors and concatenate these vectors into one collaborative instance-level guidance vector. Thus, the guidance vector contains the information from both the first and previous frames, and both the foreground and background regions.

\setlength{\intextsep}{0pt}

In order to efficiently utilize the instance-level information, we employ an attention mechanism to adjust our Collaborative Ensembler (CE). We show a detailed illustration in Fig.~\ref{fig:instance}. Inspired by SE-Nets~\cite{senet}, we leverage a fully-connected (FC) layer (we found this setting is better than using two FC layers as adopted by SE-Net) and a non-linear activation function to construct a gate for the input of each Res-Block in the CE. The gate will adjust the scale of the input feature channel-wisely.

By introducing collaborative instance-level attention, we can leverage a full scale of foreground-background information to guide the prediction further. The information with a large (instance-level) receptive field is useful to relieve local ambiguities~\cite{torralba2003contextual}, which is inevitable with a small (pixel-wise) receptive field.

\subsection{Collaborative Ensembler (CE)}

In the lower right of Fig.~\ref{fig:overview}, we design a collaborative ensembler for making large receptive fields to aggregate pixel-level and instance-level information and implicitly learn the collaborative relationship between foreground and background. 

Inspired by ResNets~\cite{resnet} and Deeplabs~\cite{deeplab,deeplabv3p}, which both have shown significant representational power in image segmentation tasks, our CE uses a downsample-upsample structure, which contains three stages of Res-Blocks~\cite{resnet} and an Atrous Spatial Pyramid Pooling (ASPP)~\cite{deeplabv3p} module. The number of Res-Blocks in Stage 1, 2, and 3 are $2$, $3$, $3$ in order. Besides, we employ dilated convolutional layers to improve the receptive fields efficiently. The dilated rates of the $3\times3$ convolutional layer of Res-Blocks in one stage are separately $1$, $2$, $4$ ( or $1$, $2$ for Stage 1). At the beginning of Stage 2 and Stage 3, the feature maps will be downsampled by the first Res-Block with a stride of 2. After these three stages, we employ an ASPP and a Decoder~\cite{deeplabv3p} module to increase the receptive fields further, upsample the scale of feature and fine-tune the prediction collaborated with the low-level backbone features.

\section{Implementation Details}
\setlength{\intextsep}{-10pt}
\begin{wrapfigure}[11]{r}{0.5\textwidth}
\center\vspace{-9mm}

\subfloat[Normal]{
\label{fig:normal_crop}
\includegraphics[width=0.426\linewidth]{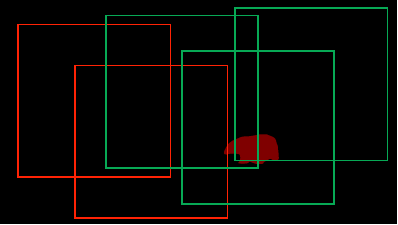}
}
\subfloat[Balanced]{
\label{fig:balanced_crop}
\includegraphics[width=0.52\linewidth]{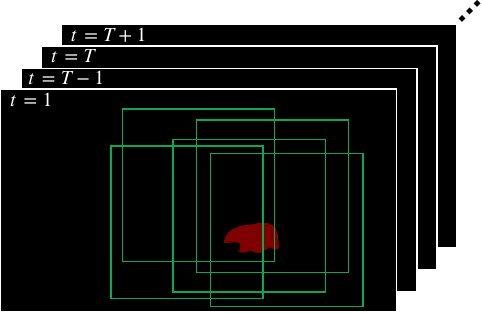}
}

\caption{When using normal random-crop, some red windows contain few or no foreground pixels. For reliving this problem, we propose balanced random-crop.}\label{fig:crop}

\end{wrapfigure}

For better convergence, we modify the random-crop augmentation and the training method in previous methods~\cite{spacetime,feelvos}.

\setlength{\intextsep}{0pt}

\noindent\textbf{Balanced Random-Crop.}
As shown in Fig.~\ref{fig:crop}, there is an apparent imbalance between the foreground and the background pixel number on VOS datasets. Such an issue usually makes the models easier to be biased to background attributes. 

In order to relieve this problem, we take a balanced random-crop scheme, which crops a sequence of frames (\ie, the first frame, the previous frame, and the current frame) by using a same cropped window and restricts the cropped region of the first frame to contain enough foreground information. The restriction method is simple yet effective. To be specific, the balanced random-crop will decide on whether the randomly cropped frame contains enough pixels from foreground objects or not. If not, the method will continually take the cropping operation until we obtain an expected one.

\noindent\textbf{Sequential Training.} \zongxin{In the training stage, FEELVOS predicts only one step in one iteration, and the guidance masks come from the ground-truth data. RGMP and STMVOS uses previous guidance information (mask or feature memory) in training, which is more consistent with the inference stage and performs better. In the evaluation stage, the previous guidance masks are always generated by the network in the previous inference steps.}

\zongxin{Following RGMP, we train the network using a sequence of consecutive frames in each SGD iteration. 
In each iteration, we randomly sample a batch of video sequences. For each video sequence, we randomly sample a frame as the reference frame and a continuous $N+1$ frames as the previous frame and current frame sequence with $N$ frames. When predicting the first frame, we use the ground-truth of the previous frame as the previous mask. When predicting the following frames, we use the latest prediction as the previous mask. 
}

\begin{figure}[t!]
    \centering
    \includegraphics[width=0.9\linewidth]{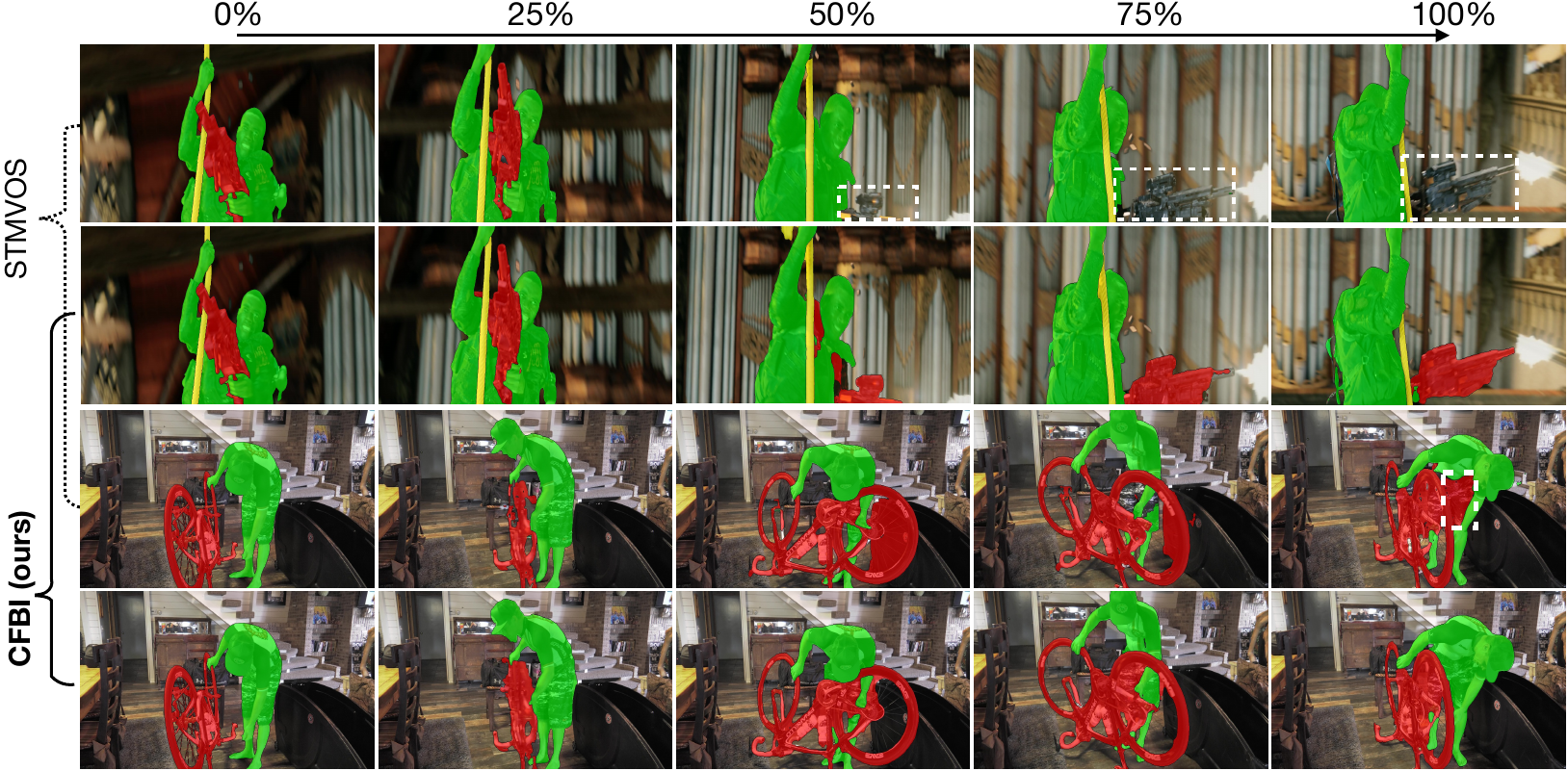}

    \caption{Qualitative comparison with STMVOS on DAVIS 2017. In the first video, STMVOS fails in tracking the gun after occlusion and blur. In the second video, STMVOS is easier to partly confuse with bicycle and person.}
    \label{fig:comparison}

\end{figure}

\noindent\textbf{Training Details.}
Following FEELVOS, we use the DeepLabv3+~\cite{deeplabv3p} architecture as the backbone for our network. However, our backbone is based on the dilated Resnet-101~\cite{deeplabv3p} instead of Xception-65~\cite{xception} for saving computational resources. We apply batch normalization (BN)~\cite{bn} in our backbone and pre-train it on ImageNet~\cite{deng2009imagenet} and COCO~\cite{coco}. The backbone is followed by one depth-wise separable convolution for extracting pixel-wise embedding with a stride of 4.

We initialize $b_B$ and $b_F$ to $0$. For the multi-local matching, we further downsample the embedding feature to a half size using bi-linear interpolation for saving GPU memory. Besides, the window sizes in our setting are $K=\{2, 4, 6, 8, 10, 12\}$. For the collaborative ensembler, we apply group normalization (GN)~\cite{gn} and gated channel transformation~\cite{gct} to improving training stability and performance when using a small batch size. For sequential training, the current sequence's length is $N=3$, which makes a better balance between computational resources and network performance.

We use the DAVIS 2017~\cite{davis2017} training set (60 videos) and the YouTube-VOS~\cite{youtubevos} training set (3471 videos) as the training data. \zongxin{We downsample all the videos to 480P resolution, which is same as the default setting in DAVIS.} We adopt SGD with a momentum of $0.9$ and apply a bootstrapped cross-entropy loss, which only considers the $15\%$ hardest pixels. During the training stage, we freeze the parameters of BN in the backbone. For the experiments on YouTube-VOS, we use a learning rate of $0.01$ for $100,000$ steps with a batch size of 4 videos (\ie, 20 frames in total) per GPU using $2$ Tesla V100 GPUs. The training time on YouTube-VOS is about 5 days. For DAVIS, we use a learning rate of $0.006$ for $50,000$ steps with a batch size of 3 videos (\ie, 15 frames in total) per GPU using $2$ GPUs.
We apply flipping, scaling, and balanced random-crop as data augmentations. The cropped window size is $465\times 465$. For the multi-scale testing, we apply the scales of $\{1.0, 1.15, 1.3, 1.5\}$ and $\{2.0, 2.15, 2.3\}$ for YouTube-VOS and DAVIS, respectively. CFBI achieves similar results in PyTorch~\cite{pytorch} and PaddlePaddle~\cite{paddlepaddle}.

\setlength{\intextsep}{-3pt}
\begin{wraptable}[28]{r}{0.55\textwidth}
	\centering\vspace{-9mm}
	\caption{The quantitative evaluation on YouTube-VOS~\cite{youtubevos}. F, S, and $^*$ separately denote fine-tuning at test time, using simulated data in the training process and performing model ensemble in evaluation. CFBI$^{MS}$ denotes using a multi-scale and flip strategy in evaluation.}\label{tab:youtubevos}
	\begin{tabular}{lccccccc}
\toprule[1.5pt]
            &  &  &  &  \multicolumn{2}{c}{Seen}  &    \multicolumn{2}{c}{Unseen} \\
\midrule[1pt]
 Methods & F & S  & Avg & $\mathcal{J}$ & $\mathcal{F}$ & $\mathcal{J}$ & $\mathcal{F}$ \\
\midrule[1pt]
\multicolumn{8}{c}{\textit{Validation 2018 Split}} \\
\midrule[1pt]
AG~\cite{agame} &   &   &  66.1  &  67.8  &  -  &  60.8  &  - \\
PReM~\cite{premvos} & \checkmark &   &  66.9  &  71.4  &  75.9  &  56.5  &  63.7 \\
BoLT~\cite{boltvos} & \checkmark &   &  71.1  &  71.6  &  -  &  64.3  &  - \\
STM$^-$~\cite{spacetime} &  &   &  68.2  &  -  &  -  &  -  &  -  \\
STM~\cite{spacetime} &  & \checkmark  &  79.4  &  79.7  &  84.2  &  72.8  &  80.9  \\
\hline
CFBI &  &   &  \textbf{81.4}  &  \textbf{81.1}  & \textbf{85.8}  & \textbf{75.3}  & \textbf{83.4}  \\
CFBI$^{MS}$ &  &   &  \textbf{82.7}  &  \textbf{82.2}  & \textbf{86.8}  & \textbf{76.9}  & \textbf{85.0}  \\
\midrule[1pt]
\multicolumn{8}{c}{\textit{Validation 2019 Split}} \\
\midrule[1pt]
CFBI &  &   &  \textbf{81.0}  &  \textbf{80.6}  & \textbf{85.1}  & \textbf{75.2}  & \textbf{83.0}  \\
CFBI$^{MS}$ &  &   &  \textbf{82.4}  &  \textbf{81.8}  & \textbf{86.1}  & \textbf{76.9}  & \textbf{84.8}  \\
\bottomrule[1.5pt]
\multicolumn{8}{c}{\textit{Testing 2019 Split}} \\
\midrule[1pt]
MST$^*$~\cite{mst} & & \checkmark &  81.7  &  80.0  &  83.3  &  \textbf{77.9}  &  85.5 \\
EMN$^*$~\cite{emn} & & \checkmark  &  81.8  &  \textbf{80.7}  &  \textbf{84.7}  &  77.3  &  84.7 \\
\hline
CFBI & & &  81.5  &  79.6  & 84.0 & 77.3  & 85.3  \\
CFBI$^{MS}$ & & &  \textbf{82.2}  &  80.4  & \textbf{84.7}  & \textbf{77.9}  & \textbf{85.7}  \\
\bottomrule[1.5pt]
\end{tabular}
\end{wraptable}

\section{Experiments}

Following the previous state-of-the-art method~\cite{spacetime},
we evaluate our method on YouTube-VOS~\cite{youtubevos}, DAVIS 2016~\cite{davis2016} and DAVIS 2017~\cite{davis2017}. For the evaluation on YouTube-VOS, we train our model on the YouTube-VOS training set~\cite{youtubevos} (3471 videos). For DAVIS, we train our model on the DAVIS-2017 training set~\cite{davis2017} (60 videos). Both DAVIS 2016 and 2017 are evaluated using an identical model trained on DAVIS 2017 for a fair comparison with the previous works~\cite{feelvos,spacetime}. Furthermore, we provide DAVIS results using both DAVIS 2017 and YouTube-VOS for training following some latest works~\cite{feelvos,spacetime}.

\setlength{\intextsep}{0pt}

The evaluation metric is the $\mathcal{J}$ score, calculated as the average IoU between the prediction and the ground truth mask, and the $\mathcal{F}$ score, calculated as an average boundary similarity measure between the boundary of the prediction and the ground truth, and their average value ($\mathcal{J}$\&$\mathcal{F}$). We evaluate our results on the official evaluation server or use the official tools.

\subsection{Compare with the State-of-the-art Methods}

\noindent \textbf{YouTube-VOS}~\cite{youtubevos} is the latest large-scale dataset for multi-object video segmentation. Compared to the popular DAVIS benchmark that consists of $120$ videos, YouTube-VOS is about 37 times larger. In detail, the dataset contains 3471 videos in the training set (65 categories), 507 videos in the validation set (additional 26 unseen categories), and 541 videos in the test set (additional 29 unseen categories). Due to the existence of unseen object categories, the YouTube-VOS validation set is much suitable for measuring the generalization ability of different methods. 

\begin{figure}[t!]
    \centering
    \includegraphics[width=0.9\linewidth]{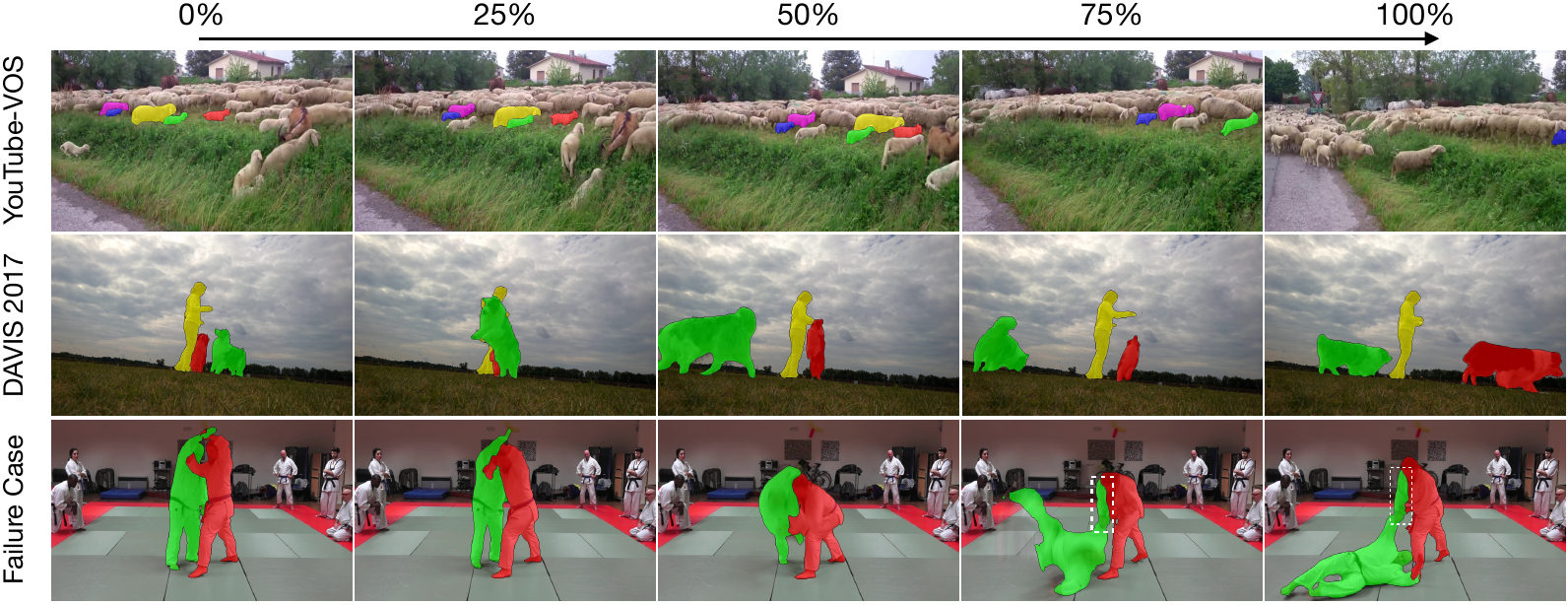}
    \caption{Qualitative results on DAVIS 2017 and YouTube-VOS. In the first video, we succeed in tracking many similar-looking sheep. In the second video, our CFBI tracks the person and the dog with a red mask after occlusion well. In the last video, CFBI fails to segment one hand of the right person (the white box). A possible reason is that the two persons are too similar and close.}
    \label{fig:quality}
\end{figure}

\begin{wraptable}[21]{r}{0.58\textwidth}
\centering
\caption{The quantitative evaluation on DAVIS 2016~\cite{davis2016} validation set. (\textbf{Y}) denotes using YouTube-VOS for training.}\label{tab:davis2016}
\begin{tabular}{l c c c c c c}
\toprule[1.5pt]
Methods  & F & S  & Avg & $\mathcal{J}$ & $\mathcal{F}$ & t/s \\
\midrule[1pt]
OSMN~\cite{osmn} &  &   &  - & 74.0  &   & 0.14 \\
PML~\cite{pml} &  &   &  77.4 & 75.5  & 79.3  & 0.28 \\
VideoMatch~\cite{videomatch} &  &  &   80.9  & 81.0  & 80.8  & 0.32 \\
RGMP$^-$~\cite{rgmp} &  &   & 68.8  & 68.6  & 68.9  & 0.14 \\
RGMP~\cite{rgmp} &  & \checkmark  & 81.8 &  81.5 & 82.0  & 0.14 \\
A-GAME~\cite{agame} (\textbf{Y}) &  &   &  82.1 & 82.2  & 82.0  & \textbf{0.07} \\
FEELVOS~\cite{feelvos} (\textbf{Y}) &  &   & 81.7  &  81.1 &  82.2 & 0.45 \\
OnAVOS~\cite{onavos}{} & \checkmark &   & 85.0  & 85.7  & 84.2  & 13 \\
PReMVOS~\cite{premvos} & \checkmark &   & 86.8  & 84.9  & 88.6  & 32.8 \\
STMVOS~\cite{spacetime} &  & \checkmark  & 86.5  & 84.8  & 88.1  & 0.16 \\
STMVOS~\cite{spacetime} (\textbf{Y}) &  & \checkmark  &  \textbf{89.3} & \textbf{88.7}  & 89.9  & 0.16 \\
\hline
CFBI  &   &   & 86.1  & 85.3 &  86.9 & 0.18 \\
CFBI (\textbf{Y})  &   &   & \textbf{89.4}  & 88.3 & \textbf{90.5}  & 0.18 \\
CFBI$^{MS}$ (\textbf{Y}) &   &   &  \textbf{90.7}  & \textbf{89.6}  & \textbf{91.7} & 9 \\
\bottomrule[1.5pt]
\end{tabular}
\end{wraptable}

As shown in Table~\ref{tab:youtubevos}, we compare our method to existing methods on both Validation 2018 and Testing 2019 splits. Without using any bells and whistles, like fine-tuning at test time~\cite{osvos,onavos} or pre-training on larger augmented simulated data~\cite{rgmp,spacetime}, our method achieves an average score of $\mathbf{81.4\%}$, which significantly outperforms all other methods in every evaluation metric. Particularly, the $81.4\%$ result is $2.0\%$ higher than the previous state-of-the-art method, STMVOS, which uses extensive simulated data from~\cite{coco,voc,cheng2014global,semantic,shi2015hierarchical} for training. Without simulated data, the performance of STMVOS will drop from $79.4\%$ to $68.2\%$. Moreover, we further boost our performance to $\mathbf{82.7\%}$ by applying a multi-scale and flip strategy during the evaluation.

We also compare our method with two of the best results on the Testing 2019 split, \ie, \textit{Rank 1} (EMN~\cite{emn}) and \textit{Rank 2} (MST~\cite{mst}) results in the 2nd Large-scale Video Object Segmentation Challenge. Without applying model ensemble, our single-model result ($\mathbf{82.2\%}$) outperforms the \textit{Rank 1} result ($81.8\%$) in the unseen and average metrics, which further demonstrates our generalization ability and effectiveness.

\noindent \textbf{DAVIS 2016}~\cite{davis2016} contains 20 videos annotated with high-quality masks each for a single target object. We compare our CFBI method with state-of-the-art methods in Table~\ref{tab:davis2016}. On the DAVIS-2016 validation set, our method trained with an additional YouTube-VOS training set achieves an average score of $\mathbf{89.4\%}$, which is slightly better than STMVOS ($89.3\%$), a method using simulated data as mentioned before. The accuracy gap between CFBI and STMVOS on DAVIS is smaller than the gap on YouTube-VOS. A possible reason is that DAVIS is too small and easy to over-fit.
Compare to a much fair baseline (\ie, FEELVOS) whose setting is same to ours, the proposed CFBI not only achieves much better accuracy ($\mathbf{89.4\%}$ \emph{vs.}\hspace{-0.8mm} $81.7\%$) but also maintains a comparable fast inference speed ($0.18s$ \emph{vs.}\hspace{-0.8mm} $0.45s$). After applying multi-scale and flip for evaluation, we can improve the performance from $\mathbf{89.4\%}$ to $\mathbf{90.1\%}$. However, this strategy will cost much more inference time ($9s$).

\setlength{\intextsep}{-3pt}
\begin{wraptable}[29]{r}{0.52\textwidth}

\caption{The quantitative evaluation on DAVIS-2017~\cite{davis2017}.}\label{tab:davis2017}
\begin{center}
\begin{tabular}{l c c c c c}
\toprule[1.5pt]
 Methods & F & S  & Avg & $\mathcal{J}$ & $\mathcal{F}$ \\
\midrule[1pt]
\multicolumn{6}{c}{\textit{Validation Split}} \\
\midrule[1pt]
OSMN~\cite{osmn} &   &   &  54.8  & 52.5  & 57.1 \\
VideoMatch~\cite{videomatch} &   &  & 62.4  & 56.5 & 68.2 \\
OnAVOS~\cite{onavos} & \checkmark  &   &  63.6  & 61.0  & 66.1 \\
RGMP~\cite{rgmp} &   & \checkmark   & 66.7 & 64.8   & 68.6 \\
A-GAME~\cite{agame} (\textbf{Y}) &   &   & 70.0  &  67.2  & 72.7 \\
FEELVOS~\cite{feelvos} (\textbf{Y}) &   &   &  71.5  &  69.1  & 74.0 \\
PReMVOS~\cite{premvos} & \checkmark  &   &  77.8  &  73.9  & 81.7 \\
STMVOS~\cite{spacetime}  &   & \checkmark  &  71.6  & 69.2  & 74.0 \\
STMVOS~\cite{spacetime} (\textbf{Y}) &   & \checkmark  &  \textbf{81.8}  & \textbf{79.2}  & 84.3 \\
\hline
CFBI  &   &   &  74.9  & 72.1  & 77.7 \\
CFBI (\textbf{Y}) &   &   &  \textbf{81.9}  & \textbf{79.1}  & \textbf{84.6} \\
CFBI$^{MS}$ (\textbf{Y}) &   &   &  \textbf{83.3}  & \textbf{80.5}  & \textbf{86.0} \\
\bottomrule[1.5pt]
\multicolumn{6}{c}{\textit{Testing Split}} \\
\midrule[1pt]
OSMN~\cite{osmn} &   &   &  41.3  & 37.7  & 44.9 \\
OnAVOS~\cite{onavos} & \checkmark  &  &  56.5  & 53.4  & 59.6 \\
RGMP~\cite{rgmp} &   &  \checkmark & 52.9 & 51.3   & 54.4 \\
FEELVOS~\cite{feelvos} (\textbf{Y}) &   &  &  57.8  &  55.2  & 60.5 \\
PReMVOS~\cite{premvos}  & \checkmark  &   &  71.6  &  67.5  & 75.7 \\
STMVOS~\cite{spacetime} (\textbf{Y}) &   & \checkmark  &  72.2  & 69.3  & 75.2 \\
\hline
CFBI (\textbf{Y})&   &  &  \textbf{74.8}  & \textbf{71.1}  & \textbf{78.5} \\
CFBI$^{MS}$ (\textbf{Y}) &   &  &  \textbf{77.5}  & \textbf{73.8}  & \textbf{81.1} \\
\bottomrule[1.5pt]
\end{tabular}
\end{center}

\end{wraptable}

\noindent \textbf{DAVIS 2017}~\cite{davis2017} is a multi-object extension of DAVIS 2016. The validation set of DAVIS 2017 consists of 59 objects in 30 videos.
Next, we evaluate the generalization ability of our model on the popular DAVIS-2017 benchmark.

As shown in Table~\ref{tab:davis2017}, our CFBI makes significantly improvement over FEELVOS ($\mathbf{81.9\%}$ \emph{vs.}\hspace{-0.8mm} $71.5\%$). Besides, our CFBI without using simulated data is slightly better than the previous state-of-the-art method, STMVOS ($\mathbf{81.9\%}$ \emph{vs.}\hspace{-0.8mm} $81.8\%$). We show some examples compared with STMVOS in Fig.~\ref{fig:comparison}. Same as previous experiments, the augmentation in evaluation can further boost the results to a higher score of $\mathbf{83.3\%}$. We also evaluate our method on the testing split of DAVIS 2017, which is much more challenging than the validation split. As shown in Table~\ref{tab:davis2017}, we significantly outperforms STMVOS ($72.2\%$) by $\textbf{2.6\%}$. By applying augmentation, we can further boost the result to $\textbf{77.5\%}$. The strong results prove that our method has the best generalization ability among the latest methods.

\noindent \textbf{Qualitative Results} We show more results of CFBI on the validation set of DAVIS 2017 ($\mathbf{81.9\%}$) and YouTube-VOS ($\mathbf{81.4\%}$) in Fig.~\ref{fig:quality}. It can be seen that CFBI is capable of producing accurate segmentation under challenging situations, such as large motion, occlusion, blur, and similar objects. In the \emph{sheep} video, CFBI succeeds in tracking five selected sheep inside a crowded flock. In the \emph{judo} video, CFBI fails to segment one hand of the right person. A possible reason is that the two persons are too similar in appearance and too close in position. Besides, their hands are with blur appearance due to the fast motion.

\subsection{Ablation Study}

\setlength{\intextsep}{-2pt}
\begin{wraptable}[15]{r}{0.4\textwidth}

\centering
\caption{Ablation of background embedding. P and I separately denote the pixel-level matching and instance-level attention. $^*$ denotes removing the foreground and background bias.}\label{tab:ablation_a}
\setlength{\tabcolsep}{6.5pt}
\begin{tabular}{l c c c c}
    \toprule[1.5pt]
         P & I & Avg & $\mathcal{J}$ & $\mathcal{F}$ \\
    \midrule[1pt]
        \checkmark  & \checkmark & 74.9 & 72.1 & 77.7 \\
    \hline
        \checkmark$^*$  & \checkmark & 72.8 & 69.5 & 76.1 \\
        \checkmark  &  & 73.0 & 69.9 & 76.0 \\
          & \checkmark & 72.3 & 69.1 & 75.4 \\
          &  & 70.9 & 68.2 & 73.6 \\
    \bottomrule[1.5pt]
\end{tabular}
\end{wraptable}

We analyze the ablation effect of each component proposed in CFBI on the DAVIS-2017 validation set. Following FEELVOS, we only use the DAVIS-2017 training set as training data for these experiments.

\noindent \textbf{Background Embedding.} As shown in Table~\ref{tab:ablation_a}, we first analyze the influence of removing the background embedding while keeping the foreground only as~\cite{feelvos,osmn}. Without any background mechanisms, the result of our method heavily drops from $74.9\%$ to $70.9\%$.
This result shows that it is significant to embed both foreground and background features collaboratively. Besides, the missing of background information in the pixel-level matching or the instance-level attention will decrease the result to $73.0\%$ or $72.3\%$ separately. 
Thus, compared to instance-level attention, the pixel-level matching performance is more sensitive to the effect of background embedding. A possible reason for this phenomenon is that the possibility of existing some background pixels similar to the foreground is higher than some background instances. Finally, we remove the foreground and background bias, $b_F$ and $b_B$, from the distance metric and the result drops to $72.8\%$, which further shows that the distance between foreground pixels and the distance between background pixels should be separately considered.

\setlength{\intextsep}{-3pt}
\begin{wraptable}[11]{r}{0.55\textwidth}

\centering
\caption{Ablation of other components.}\label{tab:ablation_b}
\begin{tabular}{l c c c c}
    \toprule[1.5pt]
          & Ablation & Avg & $\mathcal{J}$ & $\mathcal{F}$ \\
    \midrule[1pt]
        0  & Ours (CFBI) & 74.9 & 72.1 & 77.7 \\
    \hline
        1  & w/o multi-local windows & 73.8 & 70.8 & 76.8 \\
        2  & w/o sequential training & 73.3 & 70.8 & 75.7 \\
        3  & w/o collaborative ensembler & 73.3 & 70.5 & 76.1 \\
        4  & w/o balanced random-crop & 72.8 & 69.8 & 75.8 \\
        5  & w/o instance-level attention & 72.7 & 69.8 & 75.5 \\
    \hline
        6  & baseline (FEELVOS) & 68.3 & 65.6 & 70.9 \\
    \bottomrule[1.5pt]
\end{tabular}
\end{wraptable}

\noindent \textbf{Other Components.} The ablation study of other proposed components is shown in Table~\ref{tab:ablation_b}. Line 0 ($74.9\%$) is the result of proposed CFBI, and Line 6 ($68.3\%$) is our baseline method reproduced by us. Under the same setting, our CFBI significantly outperforms the baseline.

In line 1, we use only one local neighborhood window to conduct the local matching following the setting of FEELVOS, which degrades the result from $74.9\%$ to $73.8\%$. It demonstrates that our multi-local matching module is more robust and effective than the single-local matching module of FEELVOS. Notably, the computational complexity of multi-local matching dominantly depends on the biggest local window size because we use the intermediate results of the local matching of the biggest window to calculate on smaller windows.

In line 2, we replace our sequential training by using ground-truth masks instead of network predictions as the previous mask. By doing this, the performance of CFBI drops from $74.9\%$ to $73.3\%$, which shows the effectiveness of our sequential training under the same setting.

In line 3, we replace our collaborative ensembler with 4 depth-wise separable convolutional layers. This architecture is the same as the dynamic segmentation head of~\cite{feelvos}. Compared to our collaborative ensembler, the dynamic segmentation head has much smaller receptive fields and performs $1.6\%$ worse.

In line 4, we use normal random-crop instead of our balanced random-crop during the training process. In this situation, the performance drops by $2.1\%$ to $72.8\%$ as well. As expected, our balanced random-crop is successful in relieving the model form biasing to background attributes.

In line 5, we disable the use of instance-level attention as guidance information to the collaborative ensembler, which means we only use pixel-level information to guide the prediction. In this case, the result deteriorates even further to $72.7$, which proves that instance-level information can further help the segmentation with pixel-level information.

In summary, we explain the effectiveness of each proposed component of CFBI. For VOS, it is necessary to embed both foreground and background features. Besides, the model will be more robust by combining pixel-level information and instance-level information, and by using more local windows in the matching between two continuous frames. Apart from this, the proposed balanced random-crop and sequential training are useful but straightforward in improving training performance.

\section{Conclusion}

This paper proposes a novel framework for video object segmentation by introducing collaborative foreground-background integration and achieves new state-of-the-art results on three popular benchmarks. Specifically, we impose the feature embedding from the foreground target and its corresponding background to be contrastive. Moreover, we integrate both pixel-level and instance-level embeddings to make our framework robust to various object scales while keeping the network simple and fast. We hope CFBI will serve as a solid baseline and help ease the future research of VOS and related areas, such as video object tracking and interactive video editing.

\noindent \textbf{Acknowledgements.} This work is partly supported by ARC DP200100938 and ARC DECRA DE190101315.

%
%
\bibliographystyle{splncs04}
\bibliography{egbib}
\end{document}